\documentclass[10pt,twocolumn,letterpaper]{article}

\usepackage{iccv}
\usepackage{times}
\usepackage{epsfig}
\usepackage{graphicx}
\usepackage{amsmath}
\usepackage{amssymb}
\usepackage{cite}
\usepackage{mathrsfs}
\usepackage{color}
\usepackage[accsupp]{axessibility}

\usepackage[breaklinks=true,bookmarks=false]{hyperref}

\iccvfinalcopy % *** Uncomment this line for the final submission

\def\iccvPaperID{****} % *** Enter the ICCV Paper ID here

\ificcvfinal\fi

\begin{document}

%%%%%%%%% TITLE
\title{Mining Contextual Information Beyond Image for Semantic Segmentation}

\author{Zhenchao Jin$^{1}$\thanks{This work was performed while Zhenchao Jin worked as an intern at ByteDance.}, Tao Gong$^{1}$, Dongdong Yu$^{2}$, Qi Chu$^{1}$\thanks{Corresponding author.}, \\Jian Wang$^{2}$, Changhu Wang$^{2}$, Jie Shao$^{2}$\\
$^{1}$University of Science and Technology of China \quad 
$^{2}$ByteDance \\
{\tt\small\{blwx@mail., gt950513@mail., qchu@\}ustc.edu.cn }\\
{\tt\small \{yudongdong, wangjian.pierre, wangchanghu, Shaojie.mail\}@bytedance.com},
}

\maketitle
\ificcvfinal\thispagestyle{empty}\fi

%%%%%%%%% ABSTRACT
\begin{abstract}
   This paper studies the context aggregation problem in semantic image segmentation. 
   The existing researches focus on improving the pixel representations by aggregating the contextual information within individual images.
   Though impressive, these methods neglect the significance of the representations of the pixels of the corresponding class beyond the input image.
   To address this, this paper proposes to mine the contextual information beyond individual images to further augment the pixel representations.
   We first set up a feature memory module, which is updated dynamically during training, to store the dataset-level representations of various categories.
   Then, we learn class probability distribution of each pixel representation under the supervision of the ground-truth segmentation.
   At last, the representation of each pixel is augmented by aggregating the dataset-level representations based on the corresponding class probability distribution.
   Furthermore, by utilizing the stored dataset-level representations, we also propose a representation consistent learning strategy to make the classification head 
   better address intra-class compactness and inter-class dispersion.
   The proposed method could be effortlessly incorporated into existing segmentation frameworks (e.g., FCN, PSPNet, OCRNet and DeepLabV3) and brings consistent performance improvements.
   Mining contextual information beyond image allows us to report state-of-the-art performance on various benchmarks: ADE20K, LIP, Cityscapes and COCO-Stuff 
   \footnote{Our code will be available at \href{https://github.com/CharlesPikachu/mcibi}{https://github.com/CharlesPikachu/\\mcibi}.}.
\end{abstract}

\vspace{-0.45cm}
%%%%%%%%% BODY TEXT
\section{Introduction}

Semantic segmentation is a long-standing challenging task in computer vision, aiming to assign semantic labels to each pixel in an image accurately.
This task is of broad interest for potential application of autonomous driving, medical diagnosing, robot sensing, to name a few.
In recent years, deep neural networks \cite{simonyan2014very,he2016deep} have been the dominant solutions \cite{cao2019gcnet,zhao2017pyramid,chen2017deeplab,chen2017rethinking,chen2018encoder,fu2019dual,wang2018non,berman2018lovasz},
where the encoder-decoder architecture proposed in FCN \cite{long2015fully} is the cornerstone of these methods.
Specifically, these studies include adopting graphical models or cascade structure to further refine the segmentation results \cite{cheng2020cascadepsp,chen2014semantic,zheng2015conditional,cheng2020cascadepsp},
designing novel backbone networks \cite{yu2017dilated,zhang2020resnest,wang2020deep} to extract more effective feature representations,
aggregating reasonable contextual information to augment pixel representations \cite{chen2017deeplab,zhao2017pyramid,wang2018non} that is also the interest of this paper, and so on.

\begin{figure}
\centering
\includegraphics[width=0.40\textwidth]{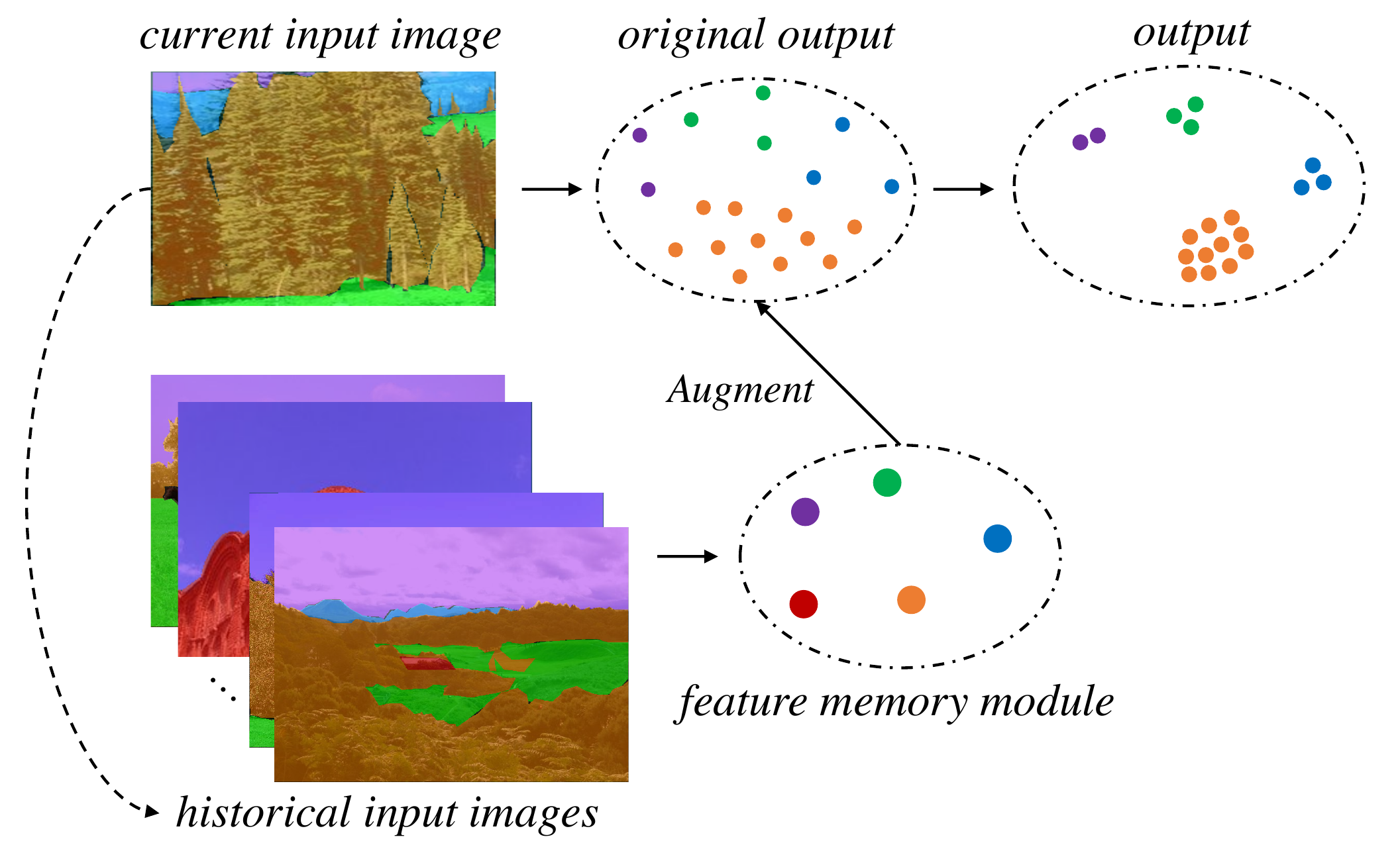}
\caption{
  Mining contextual information beyond image (MCIBI).
  The \emph{feature memory module} stores the dataset-level representations of various classes.
  The dotted line denotes that the \emph{current input image} will be added into the \emph{historical input images} after each iteration during training.
}\label{motivation}
\vspace{-0.45cm}
\end{figure}

Since there exist co-occurrent visual patterns, modeling context is universal and essential for semantic image segmentation.
Prior to this paper, the context of a pixel typically refers to a set of other pixels in the input image, \emph{e.g.}, the surrounding pixels.
Specifically, PSPNet \cite{zhao2017pyramid} utilizes pyramid spatial pooling to aggregate contextual information.
DeepLab \cite{chen2017deeplab,chen2017rethinking,chen2018encoder} exploits the contextual information by introducing the atrous spatial pyramid pooling (ASPP).
OCRNet \cite{yuan2019object} proposes to improve the representation of each pixel by weighted aggregating the object region representations.
Several recent works \cite{fu2019dual,yuan2018ocnet,huang2019ccnet,wang2018non} first calculate the relations between a pixel and its contextual pixels, 
and then aggregate the representations of the contextual pixels with higher weights for similar representations.
Apart from these methods, some studies \cite{berman2018lovasz,ke2018adaptive,zhao2019region} also find that the pixel-wise cross entropy loss 
fundamentally lacks the spatial discrimination power and thereby, propose to verify segmentation structures directly by designing context-aware optimization objectives.
Nonetheless, all of the existing approaches only model context within individual images, while discarding the potential contextual information beyond the input image.
\emph{Basically, the purpose of semantic segmentation based on deep neural networks essentially groups the representations of the pixels of the whole dataset in a non-linear embedding space.}
Consequently, to classify a pixel accurately, the semantic information of the corresponding class in the other images also makes sense.

To overcome the aforementioned limitation, this paper proposes to mine the contextual information beyond the input image so that the pixel representations could be further improved.
As illustrated in Figure \ref{motivation}, we first set up a feature memory module to store the dataset-level representations of various categories by leveraging the historical input images during training.
Then, we predict the class probability distribution of the pixel representations in the current input image, where the distribution is under the supervision of the ground-truth segmentation.
At last, each pixel representation is augmented by weighted aggregating the dataset-level representations where the weight is determined by the corresponding class probability distribution.
Furthermore, to address intra-class compactness and inter-class dispersion from the perspective of the whole dataset more explicitly, 
a representation consistent learning strategy is designed to make the classification head learn to classify both 
1) the dataset-level representations of various categories and 2) the pixel representations of the current input image.

In a nutshell, our main contributions could be summarized as:
\begin{itemize}
   \vspace{-0.15cm}
   \item To the best of our knowledge, this paper first explores the approach of mining contextual information beyond the input image to further boost the performance of semantic image segmentation. 
   \vspace{-0.15cm}
   \item A simple yet effective feature memory module is designed for storing the dataset-level representations of various categories.
   Based on this module, the contextual information beyond the input image could be aggregated by adopting the proposed dataset-level context aggregation scheme, so that
   these contexts can further enhance the representation capability of the original pixel representations.
   \vspace{-0.15cm}
   \item A novel representation consistent learning strategy is designed to better address intra-class compactness and inter-class dispersion.
   \vspace{-0.15cm}
\end{itemize}

The proposed method can be seamlessly incorporated into existing segmentation networks (\emph{e.g.}, FCN \cite{long2015fully}, PSPNet \cite{zhao2017pyramid}, OCRNet \cite{yuan2019object} and DeepLabV3 \cite{chen2017rethinking}) and brings consistent improvements.
The improvements demonstrate the effectiveness of mining contextual information beyond the input image in the semantic segmentation task.
Furthermore, introducing the dataset-level contextual information allows this paper to report state-of-the-art intersection-over-union segmentation scores on various benchmarks: ADE20K, LIP, Cityscapes and COCO-Stuff.
We expect this work to present a novel perspective for addressing the context aggregation problem in semantic image segmentation.

\section{Related Work}

\noindent \textbf{Semantic Segmentation.}
The last years have seen great development of semantic image segmentation based on deep neural networks \cite{simonyan2014very,he2016deep}.
FCN \cite{long2015fully} is the first approach to utilize fully convolutional network for semantic segmentation.
Later, many efforts \cite{chen2017deeplab,chen2017rethinking,chen2018encoder,zhao2017pyramid,cao2019gcnet,yuan2019object} have been made to further improve FCN.
To refine the coarse predictions of FCN, some researchers propose to adopt graphical models such as CRF \cite{liu2017deep,chen2014semantic,zheng2015conditional} and region growing \cite{dias2018semantic}.
CascadePSP \cite{cheng2020cascadepsp} refines the segmentation results by introducing a general cascade segmentation refinement model.
Some novel backbone networks \cite{yu2017dilated,zhang2020resnest,wang2020deep} have also been designed to extract more effective semantic information.
Feature pyramid methods \cite{kirillov2019panoptic,zhao2017pyramid,liu2015parsenet,chen2018encoder}, image pyramid methods \cite{chen2016attention,lin2017refinenet,lin2016efficient}, dilated convolutions \cite{chen2017deeplab,chen2017rethinking,chen2018encoder} 
and context-aware optimization \cite{berman2018lovasz,ke2018adaptive,zhao2019region}
are four popular paradigms to address the limited receptive field/context modeling problem in FCN.
This paper focuses on the context modeling problem in FCN.

\begin{figure*}
\centering
\includegraphics[width=0.95\textwidth]{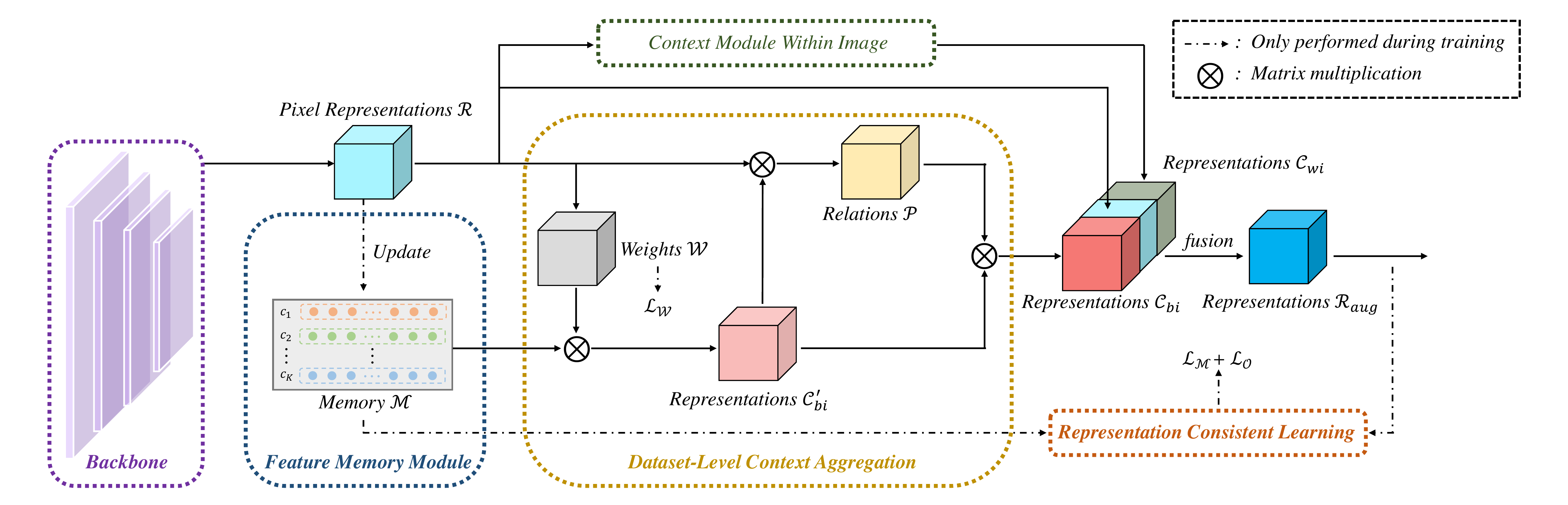}
\caption{
   Illustrating the pipeline of mining contextual information beyond the input image.
   The \emph{Context Module Within Image} denotes for applying existing context scheme within the input image (\emph{e.g.}, PPM \cite{zhao2017pyramid}, ASPP \cite{chen2017deeplab} and OCR \cite{yuan2019object}), which is an optional operation.
}\label{framework}
\vspace{-0.40cm}
\end{figure*}

\noindent \textbf{Context Scheme.}
Introducing contextual information aggregation scheme is a common practice to augment the pixel representations in semantic segmentation networks.
To obtain multi-scale contextual information, PSPNet \cite{zhao2017pyramid} performs spatial pooling at several grid scales, 
while DeepLab \cite{chen2017deeplab,chen2017rethinking,chen2018encoder} proposes to adopt pyramid dilated convolutions with different dilation rates.
Based on DeepLab, DenseASPP \cite{yang2018denseaspp} densifies the dilated rates to conver larger scale ranges.
Recently, inspired by the self-attention scheme \cite{vaswani2017attention}, 
many studies \cite{zhao2018psanet,fu2019dual,yuan2018ocnet,cao2019gcnet,wang2018non,zhu2019asymmetric} propose to augment one pixel representation by weighted aggregating all the pixel representations in the input image,
where the weights are determined by the similarities between the pixel representations.
Apart from this, some researches \cite{yuan2019object,zhang2019acfnet,li2019expectation} propose to first group the pixels into a set of regions, and then
improve the pixel representations by weighted aggregating these region representations, where the weights are the context relations between the pixel representations and region representations.
Despite impressive, all existing methods only exploit the contextual information within individual images.
Different from previous works, this paper proposes to mine more contextual information for each pixel from the whole dataset to further improve the pixel representations.

\noindent \textbf{Feature Memory Module.}
Feature memory module enables the deep neural networks to capture the effective information beyond the current input image.
It has shown power in various computer vision tasks \cite{wang2020cross,chen2020memory,zhong2019invariance,li2019memory,oh2019video,wu2019long}.
For instance, MEGA \cite{chen2020memory} adopts the memory mechanism to augment the candidate box features of key frame for video object detection.
For image search and few shot learning tasks, MNE \cite{li2019memory} proposes to leverage memory-based neighbourhood embedding to enhance a general convolutional feature.
In deep metric learning, XBM \cite{wang2020cross} designs a cross-batch memory module to mine informative examples across multiple mini-batches.
In this paper, a simple yet effective feature memory module is designed to store the dataset-level representations of various classes so that
the network can capture the contextual information beyond the input image and the representation consistent learning is able to be performed. 
To the best of our knowledge, we are the first to utilize the idea of the feature memory module in semantic image segmentation.

\vspace{-0.15cm}
\section{Methodology}
\vspace{-0.15cm}
As illustrated in Figure \ref{framework}, a backbone network is first utilized to obtain the pixel representations.
Then, the pixel representations are used to predict a weight matrix so that the dataset-level representations stored in the feature memory module can be gathered for each pixel.
Since the prediction may be inaccurate, the dataset-level representations are further refined by the pixel representations.
At last, the original pixel representations are augmented by concatenating the dataset-level representations and thereby, the augmented representations are adopted to predict the final class probability distribution of each pixel.

% \vspace{-0.15cm}
\subsection{Problem Formulation} \label{Formulation}
% \vspace{-0.15cm}
Given an input image $\mathcal{I} \in \mathbb{R}^{3 \times H \times W}$, we first project the pixels in $\mathcal{I}$ into a non-linear embedding space by a backbone network $\mathscr{B}$ as follows:
\begin{equation} \label{eq1}
   \mathcal{R} = \mathscr{B}(\mathcal{I}),
\end{equation}
where the matrix $\mathcal{R}$ of size $C \times \frac{H}{8} \times \frac{W}{8}$ stores the pixel representations of $\mathcal{I}$ and the dimension of a pixel representation is $C$.
Then, to mine the contextual information beyond the input image $\mathcal{C}_{bi}$, we have:
\begin{equation} \label{eq2}
   \mathcal{C}_{bi} = \mathscr{A}_{bi}(\mathscr{H}_1(\mathcal{R}), ~\mathcal{M}),
\end{equation}
where the dataset-level representations of various categories are stored in the feature memory module $\mathcal{M}$ and $\mathscr{A}_{bi}$ is the proposed dataset-level context aggregation scheme.
The matrix $\mathcal{C}_{bi}$ of size $C \times \frac{H}{8} \times \frac{W}{8}$ stores the dataset-level contextual information aggregated from $\mathcal{M}$.
$\mathscr{H}_1$ is a classification head used to predict the category probability distribution of the pixel representations in $\mathcal{R}$.

To integrate the proposed method into existing segmentation frameworks seamlessly, we define the self-existing context scheme of the adopted framework as $\mathscr{A}_{wi}$, and then we have:
\begin{equation} \label{eq3}
   \mathcal{C}_{wi} = \mathscr{A}_{wi}(\mathcal{R}),
\end{equation}
where $\mathcal{C}_{wi}$ is a matrix storing contextual information within the current input image.
Next, $\mathcal{R}$ is augmented as the aggregation of three parts:
\begin{equation} \label{eq4}
   \mathcal{R}_{aug} = \mathscr{F}(\mathcal{R}, ~\mathcal{C}_{bi}, ~\mathcal{C}_{wi}),
\end{equation}
where $\mathscr{F}$ is a transform function used to fuse the original representations $\mathcal{R}$, contextual representations beyond the image $\mathcal{C}_{bi}$ and contextual representations within the image $\mathcal{C}_{wi}$.
Note that, $\mathcal{C}_{wi}$ is optional according to the adopted segmentation framework. 
For instance, if the adopted framework is FCN without any contextual modules, $\mathcal{R}_{aug}$ will be calculated by $\mathscr{F}(\mathcal{R}, ~\mathcal{C}_{bi})$.
Finally, $\mathcal{R}_{aug}$ is leveraged to predict the label of each pixel in the input image:
\begin{equation} \label{eq5}
   \mathcal{O} = Upsample_{8\times}(\mathscr{H}_2 (\mathcal{R}_{aug})),
\end{equation}
where $\mathscr{H}_2$ is a classification head and $\mathcal{O}$ is a matrix of size $K \times H \times W$ storing the predicted class probability distribution.
$K$ is the number of the classes.

\subsection{Feature Memory Module} \label{FMM}
As illustrated in Figure \ref{framework}, feature memory module $\mathcal{M}$ of size $K \times C$ is introduced to store the dataset-level representations of various classes.
During training, we first randomly select one pixel representation for each category from the train set to initialize $\mathcal{M}$, where the representations are calculated by leveraging $\mathscr{B}$.
Then, the values in $\mathcal{M}$ are updated by leveraging moving average after each training iteration:
\begin{equation} \label{eq6}
   \mathcal{M}_{t} = (1 - m_{t-1}) \cdot \mathcal{M}_{t-1} + m_{t-1} \cdot \mathscr{T}(\mathcal{R}_{t-1}),
\end{equation}
where $m$ is the momentum, $t$ denotes for the current number of iterations and $\mathscr{T}$ is used to transform $\mathcal{R}$ to have the same size as $\mathcal{M}$.
For $m$, we adopt polynomial annealing policy to schedule it:
\begin{equation} \label{eq7}
   m_{t} = (1 - \frac{t}{T}) ^ p \cdot (m_{0} - \frac{m_{0}}{100}) + \frac{m_{0}}{100}, ~t \in \left[0, T\right],
\end{equation}
where $T$ is the total number of iterations of training. By default, both $p$ and $m_0$ are set as $0.9$ empirically.

To implement $\mathscr{T}$, we first setup a matrix $\mathcal{R}'$ of size $K \times C$ and initialize it by using the values in $\mathcal{M}$. 
For the convenience of presentation, we leverage the subscript $[i, j]$ or $[i, *]$ to index the element or elements of a matrix. 
$\mathcal{R}$ is upsampled and permuted as size $HW \times C$, \emph{i.e.}, $\mathcal{R}^{HW \times C}$.
Subsequently, for each category $c_k$ existing in $\mathcal{I}$, we have:
\begin{equation} \label{eq8}
   \mathcal{R}_{c_k} = \{\mathcal{R}^{HW \times C}_{[i, *]} ~|~  (\mathcal{GT}_{[i]} = c_k) \land (1 \leq i \leq HW) \},
\end{equation}
where $\mathcal{GT}$ of size $HW$ stores the ground truth category labels of $\mathcal{R}^{HW \times C}$.
$\mathcal{R}_{c_k}$ of size $N_{c_k} \times C$ stores the representations of category $c_k$ of $\mathcal{R}^{HW \times C}$.
$N_{c_k}$ is the number of pixels labeled as $c_k$ in $\mathcal{I}$. Next, we calculate the cosine similarity matrix $\mathcal{S}_{c_k}$ of size $N_{c_k}$ between $\mathcal{R}_{c_k}$ and $\mathcal{M}_{[c_k, *]}$:
\begin{equation} \label{eq9}
   \mathcal{S}_{c_k} = \frac{\mathcal{R}_{c_k} \cdot \mathcal{M}_{[c_k, *]}}{{\lVert \mathcal{R}_{c_k} \rVert}_2 \cdot {\lVert \mathcal{M}_{[c_k, *]} \rVert}_2}.
\end{equation}
Finally, the representation of $c_k$ in $\mathcal{R}'$ is updated as:
\begin{equation} \label{eq10}
   \mathcal{R}'_{[c_k, *]} = \sum^{N_{c_k}}_{i=1} \frac{1 - S_{c_k, [i]}}{\sum^{N_{c_k}}_{j=1} (1 - S_{c_k,[j]})} \cdot \mathcal{R}_{c_k, [i, *]}.
\end{equation}
The output of $\mathscr{T}$ is $\mathcal{R}'$ which has been updated by all the representations of various classes in $\mathcal{R}^{HW \times C}$.

\subsection{Dataset-Level Context Aggregation} \label{DCA}
Here, we elaborate the dataset-level context aggregation scheme $\mathscr{A}_{bi}$ to adaptively aggregate dataset-level representations stored in $\mathcal{M}$ for $\mathcal{R}$.
As demonstrated in Figure \ref{framework}, we first predict a weight matrix $\mathcal{W}$ of size $K \times \frac{H}{8} \times \frac{W}{8}$ according to $\mathcal{R}$:
\begin{equation} \label{eq11}
   \mathcal{W} = \mathscr{H}_1 (\mathcal{R}),
\end{equation}
where $\mathcal{W}$ stores the category probability distribution of each representation in $\mathcal{R}$.
The classification head $\mathscr{H}_1$ is implemented by two $1 \times 1$ convolutional layers and the $Softmax$ function.
Next, we can calculate a coarse dataset-level representation matrix $C'_{bi}$ as follows:
\begin{equation} \label{eq12}
   \mathcal{C}'_{bi} = permute(\mathcal{W}) \otimes \mathcal{M},
\end{equation}
where $\mathcal{C}'_{bi}$ of size $\frac{HW}{64} \times C$ stores the aggregated dataset-level representations for the pixel representations in $\mathcal{R}$.
$permute(\mathcal{W})$ is used to make $\mathcal{W}$ have size of $\frac{HW}{64} \times K$ and $\otimes$ denotes for the matrix multiplication.
Since $\mathscr{H}_1$ only leverages $\mathcal{R}$ to predict $\mathcal{W}$, the pixel representations may be misclassified.
To address this, we calculate the relations between $\mathcal{R}$ and $\mathcal{C}'_{bi}$ so that we can obtain a position confidence weight to further refine $\mathcal{C}'_{bi}$.
Specifically, we first calculate the relations $\mathcal{P}$ as follows:
\begin{equation} \label{eq13}
   \mathcal{P} = Softmax(\frac{g_q(permute(\mathcal{R})) \otimes g_k(\mathcal{C}^{'}_{bi})^T}{\sqrt{\frac{C}{2}}}),
\end{equation}
where $permute$ is adopted to let $\mathcal{R}$ have size of $\frac{HW}{64} \times C$.
Then, $\mathcal{C}^{'}_{bi}$ is refined as follows:
\begin{equation} \label{eq14}
   \mathcal{C}_{bi} = permute(g_{o}(\mathcal{P} \otimes g_v(\mathcal{C}_{bi}^{'}))),
\end{equation}
where $g_q$, $g_k$, $g_v$ and $g_o$ are introduced to adjust the dimension of each pixel representation, implemented by a $1 \times 1$ convolutional layer.
$permute$ is used to let the output have size of $C \times \frac{H}{8} \times \frac{W}{8}$.

\subsection{Representation Consistent Learning} \label{RCL}
Segmentation model based on deep neural networks essentially groups the representations of the pixels of the whole dataset in a non-linear embedding space,
while the deep architectures are typically trained by a mini-batch of the dataset at each iteration.
Although this can prevent the models from overfitting in a sense, the inconsistency makes the network lack the ability of addressing intra-class compactness and inter-class dispersion from the perspective of the whole dataset.
To alleviate this problem, we propose a representation consistent learning strategy.

Specifically, during training, $\mathscr{H}_2$ is also leveraged to predict the labels of the dataset-level representations in $\mathcal{M}$:
\begin{equation} \label{eq15}
   \mathcal{O}^{\mathcal{M}} = \mathscr{H}_2 (reshape(\mathcal{M})),
\end{equation}
where $reshape$ is utilized to make $\mathcal{M}$ have size of $K \times C \times 1 \times 1$. 
The classification head $\mathscr{H}_2$ is implemented by two $1 \times 1$ convolutional layers and the $Softmax$ function.
$\mathcal{O}^{\mathcal{M}}$ stores the predicted category probability distribution of each dataset-level representation in $\mathcal{M}$.
It is clear that each representation in $\mathcal{M}$ is  essentially a composite vector of the pixel representations of the same category of the whole dataset.
Therefore, sharing the classification head in predicting $\mathcal{O}$ and $\mathcal{O}^{\mathcal{M}}$ could make $\mathscr{H}_2$ not only 
1) address the categorization ability of the pixels within individual images, but also 
2) learn to address intra-class compactness and inter-class dispersion from the perspective of the whole dataset more explicitly.

\subsection{Loss Function} \label{LF}
A multi-task loss function of $\mathcal{W}$, $\mathcal{O}^{\mathcal{M}}$ and $\mathcal{O}$ is used to jointly optimize the model parameters.
Specifically, the objective of $\mathcal{W}$ is defined as follows:
\begin{equation} \label{eq16}
   \mathcal{L}_{\mathcal{W}} = \frac{1}{H \times W} \sum_{i, j} L_{ce} (\mathcal{W}^{K \times H \times W}_{[*, i,j]}, ~\S (\mathcal{GT}_{[ij]})),
\end{equation}
where $\S$ stands for converting the ground truth category label into one-hot format and $\mathcal{W}^{K \times H \times W}$ is the output of $Upsample_{8 \times}(\mathcal{W})$.
$L_{ce}$ denotes for the \emph{cross entropy loss} and $\sum_{i, j}$ denotes that the summation is calculated over all locations on the input image $\mathcal{I}$.
For $\mathcal{O}^{\mathcal{M}}$, the loss function is formulated as follows:
\begin{equation} \label{eq17}
   \mathcal{L}_{{\mathcal{M}}} = \frac{1}{K} \sum^{K}_{k=1} L_{ce}(\mathcal{O}^{\mathcal{M}}_{[*, c_k, 1, 1]},~\S(c_k)).
\end{equation}
To make $\mathcal{O}$ contain the accurate category probability distribution of each pixel, we define the loss function of $\mathcal{O}$ as follows:
\begin{equation} \label{eq18}
   \mathcal{L}_{\mathcal{O}} = \frac{1}{H \times W} \sum_{i, j} L_{ce} (\mathcal{O}_{[*, i, j]}, ~\S (\mathcal{GT}_{[ij]})).
\end{equation}
Finally, we formulate the multi-task loss function $\mathcal{L}$ as:
\begin{equation} \label{eq19}
   \mathcal{L} = \alpha \mathcal{L}_{\mathcal{W}} + \beta \mathcal{L}_{\mathcal{M}} + \mathcal{L}_{\mathcal{O}},
\end{equation}
where $\alpha$ and $\beta$ are the hyper-parameters to balance the three losses. 
We empirically set $\alpha=0.4$ and $\beta = 1$ by default.
With this joint loss function, the model parameters are learned jointly through back propagation.
Note that the values in $\mathcal{M}$ are not learnable through back propagation.

\section{Experiments}
\subsection{Experimental Setup}
\noindent \textbf{Datasets.}
We conduct experiments on four widely-used semantic segmentation benchmarks:
\begin{itemize}
   \item \textbf{ADE20K} \cite{zhou2017scene} is a challenging scene parsing dataset that contains
   150 classes and diverse scenes with 1,038 image-level labels.
   The training, validation, and test sets consist of 20K, 2K, 3K images, respectively.
   \vspace{-0.15cm}
   \item \textbf{LIP} \cite{gong2017look} is tasked for single human parsing. There are 19 semantic classes and 1 background class.
   The dataset is divided into 30K/10K/10K images for training, validation and testing.
   \vspace{-0.15cm}
   \item \textbf{Cityscapes} \cite{cordts2016cityscapes} densely annotates 19 object categories in urban scene understanding images.
   This dataset consists of 5,000 finely annotated images, out of which 2,975 are available for training, 500 for validation and 
   1,524 for testing.
   \vspace{-0.15cm}
   \item \textbf{COCO-Stuff} \cite{caesar2018coco} is a challenging scene parsing dataset that provides rich annotations for 
   91 thing classes and 91 stuff classes. This paper adopts the latest version which includes all 164K images 
   (\emph{i.e.}, 118K for training, 5K for validation, 20K for test-dev and 20K for test-challenge) from MS COCO \cite{lin2014microsoft}.
\end{itemize}

\noindent \textbf{Training Settings.} 
Our backbone network is initialized by the weights pre-trained on ImageNet \cite{deng2009imagenet}, while the remaining layers are randomly initialized. 
Following previous work \cite{chen2017deeplab}, the learning rate is decayed according to the ``poly'' learning rate policy with factor $(1 - \frac{iter}{total\_iter})^{0.9}$.
Synchronized batch normalization implemented by pytorch is enabled during training.
For data augmentaton, we adopt random scaling, horizontal flipping and color jitter.
More specifically, the training settings for different datasets are listed as follows:
\begin{itemize}

   \item \textbf{ADE20K:} 
   We set the initial learning rate as $0.01$,
   weight decay as $0.0005$, 
   crop size as $512 \times 512$, 
   batch size as $16$ and 
   training epochs as $130$ by default.
   \vspace{-0.15cm}

   \item \textbf{LIP:} 
   The initial learning rate is set as $0.01$ and the weight decay is $0.0005$. 
   We set the crop size of the input image as $473 \times 473$ and batch size as $32$ by default.
   Besides, if not specified, the networks are fine-tuned for $150$ epochs on the train set.
   \vspace{-0.15cm}

   \item \textbf{Cityscapes:}    
   The initial learning rate is set as $0.01$ and the weight decay is $0.0005$.
   We randomly crop out high-resolution patches $512 \times 1024$ from the original images as the inputs for training.
   The batch size and training epochs are set as $8$ and $220$, respectively.
   \vspace{-0.15cm}

   \item \textbf{COCO-Stuff:} 
   We set the initial learning rate as $0.01$,
   weight decay as $0.0005$, 
   crop size as $512 \times 512$, 
   batch size as $16$ and 
   training epochs as $30$ by default.
   \vspace{-0.15cm}

\end{itemize}

\noindent \textbf{Inference Settings.} 
For ADE20K, COCO-Stuff and LIP, the size of the input image during testing is the same as the size of the input image during training.
And for Cityscapes, the input image is resized to have its shorter side being $1024$ pixels. 
By default, no tricks (\emph{e.g.}, multi-scale with flipping testing) will be used during testing.

\noindent \textbf{Evaluation Metrics.} 
Following the standard setting, mean intersection-over-union (mIoU) is adopted for evaluation.

\noindent \textbf{Reproducibility.} 
Our method is implemented in PyTorch ($version \geq 1.3$) \cite{paszke2019pytorch}
and trained on 8 NVIDIA Tesla V100 GPUs with a 32 GB memory per-card.
And all the testing procedures are performed on a single NVIDIA Tesla V100 GPU.
To provide full details of our framework, our code will be made publicly available.

% table1---------------------------------------------------------------------
\begin{table}[t]
\centering
\caption{
   Ablation study on the validation set of ADE20K about the form of the transform function in Equation \ref{eq4}. 
   FCN \cite{long2015fully} is adopted as the \emph{Baseline} framework.
}\label{table1}
\resizebox{.42\textwidth}{!}{
\begin{tabular}{c|c|c|c}
   \hline
   \hline
   Method                              &Backbone      &$\mathscr{F}$           &mIoU             \\
   \hline
   \emph{Baseline}                     &ResNet-50     &-                       &36.96            \\
   \hline
   \emph{Baseline}+MCIBI (\emph{ours}) &ResNet-50     &\emph{add}              &40.99            \\
   \emph{Baseline}+MCIBI (\emph{ours}) &ResNet-50     &\emph{weighted add}     &41.17            \\
   \emph{Baseline}+MCIBI (\emph{ours}) &ResNet-50     &\emph{concatenation}    &\textbf{42.18}   \\
   \hline
   \hline
\end{tabular}}
\end{table}
% table1---------------------------------------------------------------------

% table2---------------------------------------------------------------------
\begin{table}[t]
\centering
\caption{
   Ablation study on the validation set of ADE20K about the proposed feature memory module.
   ``Clustering'' denotes that each dataset-level representation of the feature memory module is initialized by clustering of the pixel representations of corresponding category.
   ``RCL'' stands for representation consistent learning.
}\label{table2}
\resizebox{.35\textwidth}{!}{
\begin{tabular}{cccc|c}
   \hline
   \hline
   Cosine                      &Poly          &Clustering    &RCL                     &mIoU             \\
   \hline
   \checkmark                  &              &              &                        &41.77            \\
                               &\checkmark    &              &                        &41.95            \\
   \checkmark                  &\checkmark    &              &                        &42.18            \\
   \checkmark                  &\checkmark    &\checkmark    &                        &42.20            \\
   \checkmark                  &\checkmark    &              &\checkmark              &\textbf{42.84}   \\
   \checkmark                  &\checkmark    &\checkmark    &\checkmark              &42.75            \\
   \hline
   \hline
\end{tabular}}
\vspace{-0.45cm}
\end{table}
% table2---------------------------------------------------------------------

\subsection{Ablation Study} \label{ablationstudy}
\noindent \textbf{Transform function in Equation \ref{eq4}.}
Table \ref{table1} shows three fusion methods to fuse $\mathcal{R}$ and $C_{bi}$.
\emph{add} denotes for element-wise adding $\mathcal{R}$ and $C_{bi}$, which brings $4.03\%$ mIoU improvements.
\emph{weighted add} means adding $\mathcal{R}$ and $C_{bi}$ weightedly where the weights are predicted by a $1 \times 1$ convolutional layer.
It achieves $4.21\%$ mIoU improvements. 
\emph{concatenation}, which stands for concatenating $\mathcal{R}$ and $C_{bi}$, is the best choice to implement $\mathscr{F}$. It is $5.22\%$ mIoU higher than \emph{Baseline}.
These results together indicate that mining contextual information beyond the input image could effectively improve the pixel representations so that the models could classify the pixels more accurately.

\noindent \textbf{Update Strategy of $\mathcal{M}$.}
The feature memory module $\mathcal{M}$ is updated by employing moving average, where the current representations of the same category are combined by calculating the cosine similarity and the momentum is adjusted by using polynomial annealing policy.
To show the effectiveness, we remove the step of calculating the cosine similarity and leveraging polynomial annealing policy, respectively.
As indicated in Table \ref{table2}, applying cosine similarity and poly policy could bring $0.41\%$ and $0.23\%$ mIoU improvements, respectively.

% table3---------------------------------------------------------------------
\begin{table}[t]
\centering
\caption{
   Complexity comparison with existing context schemes. 
   The feature map of size $[1 \times 2048 \times 128 \times 128]$ is used to evaluate their complexity during inference.
   All the numbers are obtained on a single NVIDIA Tesla V100 GPU with CUDA 11.0 and the smaller, the better.
}\label{table3}
\resizebox{.45\textwidth}{!}{
\begin{tabular}{c|c|c|c}
   \hline
   \hline
   Method                                              &Parameters      &FLOPs            &Time                  \\
   \hline
   ASPP \cite{chen2017rethinking} (\emph{our impl.})   &42.21M          &674.47G          &101.44ms              \\
   PPM \cite{zhao2017pyramid} (\emph{our impl.})       &23.07M          &309.45G          &29.57ms               \\
   CCNet \cite{huang2019ccnet} (\emph{our impl.})      &23.92M          &397.38G          &56.90ms               \\
   DANet \cite{fu2019dual} (\emph{our impl.})          &23.92M          &392.02G          &62.64ms               \\
   ANN \cite{zhu2019asymmetric} (\emph{our impl.})     &20.32M          &335.24G          &49.66ms               \\
   DNL \cite{yin2020disentangled} (\emph{our impl.})   &24.12M          &395.25G          &68.62ms               \\
   APCNet \cite{he2019adaptive} (\emph{our impl.})     &30.46M          &413.12G          &54.20ms               \\
   \hline
   DCA (\emph{ours})                                   &14.82M          &242.80G          &47.64ms               \\
   ASPP+DCA (\emph{ours})                              &45.63M          &730.56G          &125.03ms               \\
   \hline
   \hline
\end{tabular}}
\end{table}
% table3---------------------------------------------------------------------

% table4---------------------------------------------------------------------
\begin{table}[t]
\centering
\caption{
   Generalization ability of our feature memory module.
}\label{table4}
\resizebox{.42\textwidth}{!}{
\begin{tabular}{c|c|c|c}
   \hline
   \hline
   Method                                      &Train Set           &Test Set         &mIoU        \\
   \hline
   FCN \cite{long2015fully}                    &LIP train set       &LIP val set      &48.63       \\
   DeepLabV3 \cite{chen2017rethinking}         &LIP train set       &LIP val set      &52.35       \\
   FCN \cite{long2015fully}                    &LIP train set       &CIHP val set     &27.20       \\
   DeepLabV3 \cite{chen2017rethinking}         &LIP train set       &CIHP val set     &27.02       \\
   \hline
   FCN+MCIBI (\emph{ours})                     &LIP train set       &LIP val set      &\textbf{51.07}       \\
   DeepLabV3+MCIBI (\emph{ours})               &LIP train set       &LIP val set      &\textbf{53.52}       \\
   FCN+MCIBI (\emph{ours})                     &LIP train set       &CIHP val set     &\textbf{27.57}       \\
   DeepLabV3+MCIBI (\emph{ours})               &LIP train set       &CIHP val set     &\textbf{27.89}       \\
   \hline
   \hline
\end{tabular}}
\vspace{-0.45cm}
\end{table}
% table4---------------------------------------------------------------------

\noindent \textbf{Initialization Method of $\mathcal{M}$.}
By default, each dataset-level representation in $\mathcal{M}$ is initialized by randomly selecting a pixel representation of corresponding category.
To test the impact of the initialization method on segmentation performance, we also attempt to initialize $\mathcal{M}$ by cosine similarity clustering.
Table \ref{table2} shows the result, which is very comparable with random initialization ($42.18\%$ \emph{v.s.} $42.20\%$).
This result indicates that the proposed update strategy could make $\mathcal{M}$ converge well, without relying on a strict initialization, which shows the robustness of our method.

\noindent \textbf{Representation Consistent Learning.}
The representation consistent learning strategy (RCL) is designed to make the network learn to address intra-class compactness and inter-class dispersion from the perspective of the whole dataset more explicitly.
As illustrated in Table \ref{table2}, we can see that RCL brings $0.66\%$ improvements on mIoU,
which demonstrates the effectiveness of our method.

% table5---------------------------------------------------------------------
\begin{table*}[t]
\centering
\caption{
   The improvements of segmentation performance achieved on different benchmarks when integrating the proposed mining contextual information beyond image (MCIBI)
   into the existing segmentation frameworks. All the results here are obtained under single-scale testing without any tricks.
}\label{table5}
\resizebox{.9\textwidth}{!}{
\begin{tabular}{c|c|c|c|c|c|c}
   \hline
   \hline
   Method                                  &Backbone      &Stride     &ADE20K (\emph{train}/\emph{val})       &Cityscapes (\emph{train}/\emph{val})      &LIP (\emph{train}/\emph{val})       &COCO-Stuff (\emph{train}/\emph{val}) \\
   \hline
   FCN \cite{long2015fully}                &ResNet-50     &$8\times$  &36.96                                  &75.16                                     &48.63                               &35.11             \\
   FCN+MCIBI (\emph{ours})                 &ResNet-50     &$8\times$  &\textbf{42.84}                         &\textbf{78.50}                            &\textbf{51.07}                      &\textbf{39.95}    \\
   \hline
   PSPNet \cite{zhao2017pyramid}           &ResNet-50     &$8\times$  &42.64                                  &79.05                                     &51.94                               &41.01             \\
   PSPNet+MCIBI (\emph{ours})              &ResNet-50     &$8\times$  &\textbf{43.77}                         &\textbf{79.90}                            &\textbf{52.84}                      &\textbf{41.80}    \\
   \hline
   OCRNet \cite{yuan2019object}            &ResNet-50     &$8\times$  &42.47                                  &79.40                                     &52.25                               &40.57             \\
   OCRNet+MCIBI (\emph{ours})              &ResNet-50     &$8\times$  &\textbf{43.34}                         &\textbf{79.98}                            &\textbf{52.80}                      &\textbf{41.10}    \\
   \hline
   DeepLabV3 \cite{chen2017rethinking}     &ResNet-50     &$8\times$  &43.19                                  &80.57                                     &52.35                               &41.86             \\
   DeepLabV3+MCIBI (\emph{ours})           &ResNet-50     &$8\times$  &\textbf{44.34}                         &\textbf{81.14}                            &\textbf{53.52}                      &\textbf{42.08}    \\
   \hline
   \hline
\end{tabular}}
\vspace{-0.3cm}
\end{table*}
% table5---------------------------------------------------------------------

% table6---------------------------------------------------------------------
\begin{table}[t]
\centering
\caption{
   Comparison with the state-of-the-art on ADE20K (val).
   Multi-scale and flipping testing is adopted for fair comparison.
}\label{table6}
\resizebox{.45\textwidth}{!}{
\begin{tabular}{c|c|c|c}
   \hline
   \hline
   Method                                           &Backbone      &Stride                  &mIoU                  \\
   \hline
   CCNet \cite{huang2019ccnet}                      &ResNet-101    &$8\times$               &45.76                 \\
   OCNet \cite{yuan2018ocnet}                       &ResNet-101    &$8\times$               &45.45                 \\
   ACNet \cite{fu2019adaptive}                      &ResNet-101    &$8\times$               &45.90                \\
   DMNet \cite{he2019dynamic}                       &ResNet-101    &$8\times$               &45.50                 \\
   EncNet \cite{zhang2018context}                   &ResNet-101    &$8\times$               &44.65                 \\
   PSPNet \cite{zhao2017pyramid}                    &ResNet-101    &$8\times$               &43.29                 \\
   PSANet \cite{zhao2018psanet}                     &ResNet-101    &$8\times$               &43.77                 \\
   APCNet \cite{he2019adaptive}                     &ResNet-101    &$8\times$               &45.38                 \\
   OCRNet \cite{yuan2019object}                     &ResNet-101    &$8\times$               &45.28                 \\
   PSPNet \cite{zhao2017pyramid}                    &ResNet-269    &$8\times$               &44.94                 \\
   OCRNet \cite{yuan2019object}                     &HRNetV2-W48   &$4\times$               &45.66                 \\
   DeepLabV3 \cite{chen2017rethinking}              &ResNeSt-50    &$8\times$               &45.12                 \\
   DeepLabV3 \cite{chen2017rethinking}              &ResNeSt-101   &$8\times$               &46.91                 \\
   SETR-MLA \cite{zheng2021rethinking}              &ViT-Large     &$16\times$              &50.28                 \\
   \hline
   DeepLabV3+MCIBI (\emph{ours})                    &ResNet-50     &$8\times$               &45.95                 \\
   DeepLabV3+MCIBI (\emph{ours})                    &ResNet-101    &$8\times$               &47.22                 \\
   DeepLabV3+MCIBI (\emph{ours})                    &ResNeSt-101   &$8\times$               &47.36                 \\
   DeepLabV3+MCIBI (\emph{ours})                    &ViT-Large     &$16\times$              &\textbf{50.80}        \\
   \hline
   \hline
\end{tabular}}
\vspace{-0.40cm}
\end{table}
% table6---------------------------------------------------------------------

\noindent \textbf{Complexity Comparison.} 
Table \ref{table3} shows the efficiency of our dataset-level context aggregation (DCA) scheme.
Compared to the existing context schemes, our DCA requires the least parameters and the least computation complexity, which indicates that the proposed context module is light.
Since our context scheme is complementary to the existing context schemes, this result well demonstrates that the model complexity is still tolerable after integrating the proposed module into existing segmentation framework.
For instance, after integrating DCA into ASPP, the parameters, FLOPs and inference time is just increased by $2.5$M, $40.94$G and $23.59$ms, respectively.

\noindent \textbf{Generalization Ability.} 
Since the feature memory module stores the dataset-level contextual information to augment the pixel representation, this module may cause the segmentation framework to overfit on the current dataset.
To dispel this worry, we train the base model and the proposed method on the train set of LIP, and then test them on the validation set of both LIP and CIHP \cite{gong2018instance}, where CIHP is a much more challenging multi-human parsing dataset with the same categories as LIP.
As demonstrated in Table \ref{table4}, we can see that the dataset-level representations of LIP benefit the model on both LIP ($51.07\%$ \emph{v.s.} $48.63\%$ and $53.52\%$ \emph{v.s.} $52.35\%$) and CIHP ($27.57\%$ \emph{v.s.} $27.20\%$ and $27.89\%$ \emph{v.s.} $27.02\%$).
This result well proves that our method owns the generalization ability.

\noindent \textbf{Integrated into Various Segmentation Frameworks.} 
As illustrated in Table \ref{table5}, we can see that our approach improves the performance of base frameworks (\emph{i.e.}, FCN \cite{long2015fully}, PSPNet \cite{zhao2017pyramid}, OCRNet \cite{yuan2019object} and DeepLabV3 \cite{chen2017rethinking}) by solid margins.
For instance, on ADE20K, mining contextual information beyond image (MCIBI) brings $5.88\%$ mIoU improvements to a simple FCN framework.
And for stronger baselines (\emph{i.e.}, PSPNet, OCRNet and DeepLabV3), the performance still gains about $1\%$ mIoU improvements.
These results indicate that our method could be seamlessly incorporated into existing segmentation frameworks and brings consistent improvements.

\begin{figure}
\centering
\includegraphics[width=0.45\textwidth]{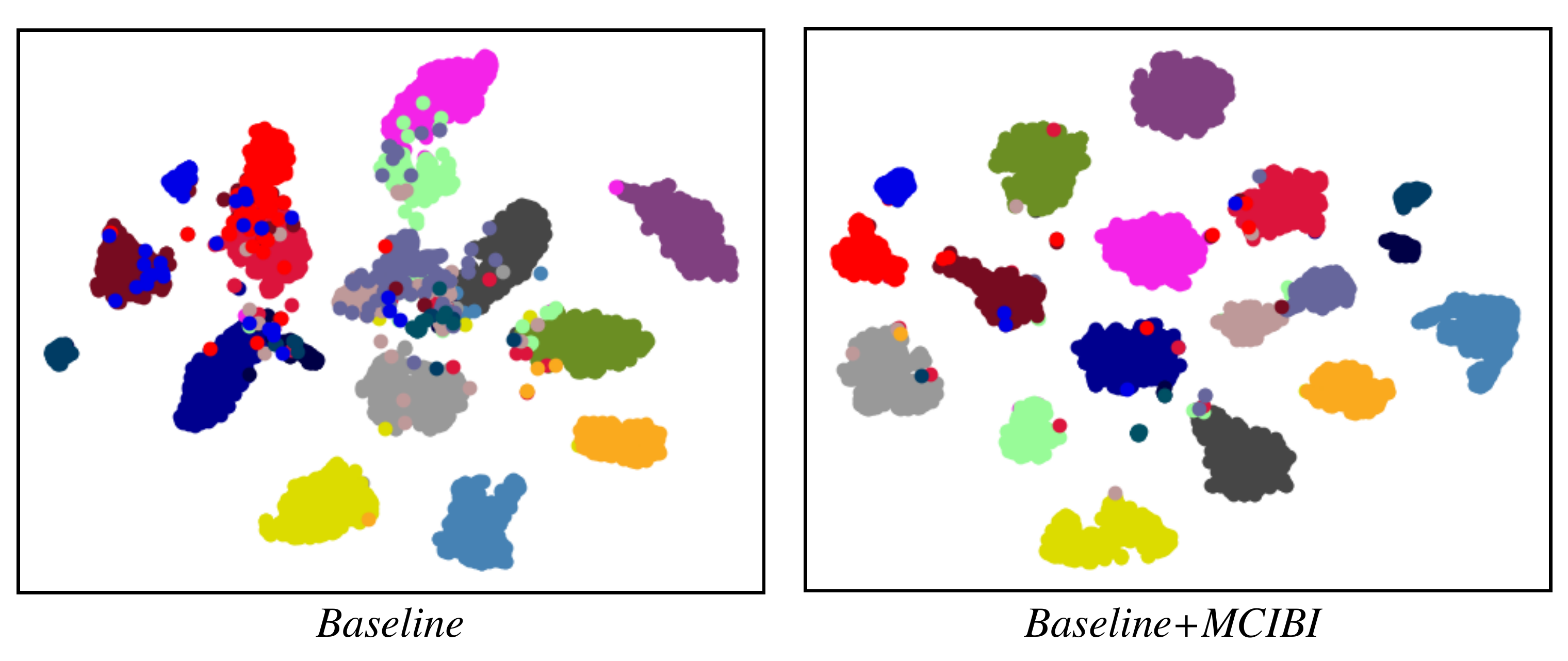}
\caption{
   t-SNE visualization on Cityscapes val.
   A point stands for a composite vector of the pixel representations of the same category within an image.
   The official recommended color table is used to color the learned features according to their category labels.
   Obviously, mining contextual information beyond image (MCIBI) can beget a more well-structured semantic feature space.
}\label{visualization}
\vspace{-0.40cm}
\end{figure}

\begin{figure*}
\centering
\includegraphics[width=1.0\textwidth]{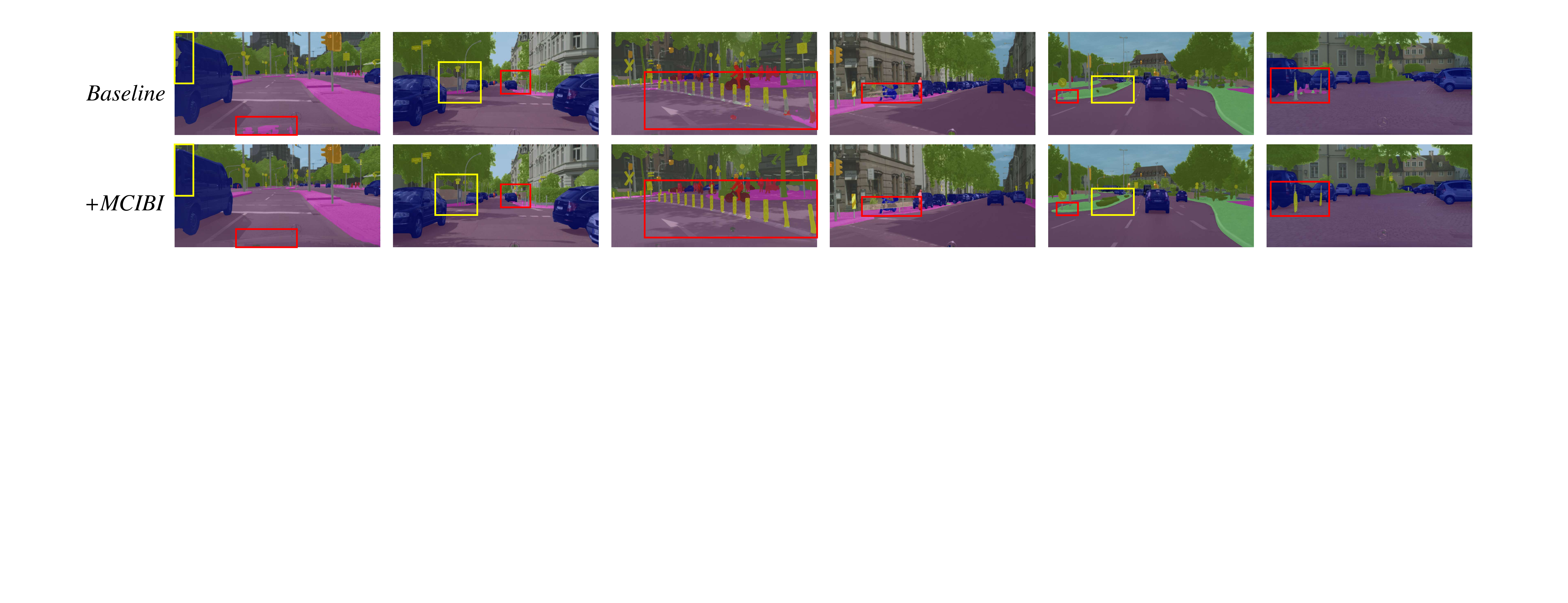}
\caption{
   Visual comparisons between FCN (\emph{baseline}) and FCN + MCIBI (\emph{ours}) on the validation set of Cityscapes.
}\label{compare}
\vspace{-0.30cm}
\end{figure*}

\noindent \textbf{Visualization of Learned Features.} 
As illustrated in Figure \ref{visualization}, we visualize the pixel representations learned by the baseline model (left) and our approach (right), respectively.
As seen, after applying our method, the spatial distribution of the learned features are well improved, \emph{i.e.}, the pixel representations with the same category are more compact and the features of various classes are well separated.
This result well demonstrates that the dataset-level representations can help augment the pixel representations by improving the spatial distribution of the original output in the non-linear embedding space so that the network can classify the pixels more smoothly and accurately.
More specifically, Figure \ref{compare} has illustrated some qualitative results of our method compared to the baseline model.

% table7---------------------------------------------------------------------
\begin{table}[t]
\centering
\caption{
   Segmentation results on the validation set of LIP.
   Flipping testing is leveraged for fair comparison.
}\label{table7}
\resizebox{.45\textwidth}{!}{
\begin{tabular}{c|c|c|c}
   \hline
   \hline
   Method                                           &Backbone      &Stride       &mIoU            \\
   \hline
   DeepLab \cite{chen2017deeplab}                   &ResNet-101    &-            &44.80           \\
   CE2P \cite{ruan2019devil}                        &ResNet-101    &$16\times$   &53.10           \\
   DeepLabV3 \cite{chen2017rethinking}              &ResNet-101    &$8\times$    &54.58           \\
   OCNet \cite{yuan2018ocnet}                       &ResNet-101    &$8\times$    &54.72           \\
   CCNet \cite{huang2019ccnet}                      &ResNet-101    &$8\times$    &55.47           \\
   OCRNet \cite{yuan2019object}                     &ResNet-101    &$8\times$    &55.60           \\
   HRNet \cite{wang2020deep}                        &HRNetV2-W48   &$4\times$    &55.90           \\
   OCRNet \cite{yuan2019object}                     &HRNetV2-W48   &$4\times$    &56.65           \\
   \hline
   DeepLabV3+MCIBI (\emph{ours})                    &ResNet-50     &$8\times$    &54.08           \\
   DeepLabV3+MCIBI (\emph{ours})                    &ResNet-101    &$8\times$    &55.42           \\
   DeepLabV3+MCIBI (\emph{ours})                    &ResNeSt-101   &$8\times$    &56.34           \\
   DeepLabV3+MCIBI (\emph{ours})                    &HRNetV2-W48   &$4\times$    &\textbf{56.99}  \\
   \hline
   \hline
\end{tabular}}
\vspace{-0.40cm}
\end{table}
% table7---------------------------------------------------------------------

% vis---------------------------------------------------------------------
% vis---------------------------------------------------------------------

% compare---------------------------------------------------------------------
% compare---------------------------------------------------------------------

\subsection{Comparison with State-of-the-Art}
\noindent \textbf{Results on ADE20K.}
Table \ref{table6} compares the segmentation results on the validation set of ADE20K.
Under ResNet-101, ACNet \cite{fu2019adaptive} achieves $45.90\%$ mIoU through combining richer local and global contexts, which is the previous best method.
By using ResNet-50, our method has achieved a very completive result of $45.95\%$ mIoU.
With the same backbone, our approach achieves superior mIoU of $47.22\%$ with a significant margin over ACNet.
Prior to this paper, SETR-MLA \cite{zheng2021rethinking} achieves the state-of-the-art with $50.28\%$ mIoU by adopting stronger backbone ViT-Large.
Based on this, our method DeepLabV3+MCIBI with ViT-Large achieves a new state-of-the-art with mIoU hitting $50.80\%$.
As is known, ADE20K is very challenging due to its various image sizes, a great deal of semantic categories and the gap between its training and validation set.
Therefore, these results can strongly demonstrate the significance of mining contextual information beyond the input image.

\noindent \textbf{Results on LIP.}
Table \ref{table7} has shown the comparison results on the validation set of LIP.
It can be seen that our DeepLabV3+MCIBI using ResNeSt101 yields an mIoU of $56.34\%$, which is among the state-of-the-art.
Furthermore, our method achieves a new state-of-the-art with mIoU hitting $56.99\%$ based on the high-resolution network HRNetV2-W48 that is more robust than ResNeSt in semantic image segmentation.
This is particularly impressive considering the fact that human parsing is a more challenging fine-grained semantic segmentation task.

\noindent \textbf{Results on Cityscapes.}
Table \ref{table8} demonstrates the comparison results on the test set of Cityscapes.
The previous state-of-the-art method OCRNet with HRNetV2-W48 achieves $82.40\%$ mIoU.
Thus, we also integrate the proposed MCIBI into HRNetV2-W48 so that we could report a new state-of-the-art performance of $82.55\%$ mIoU.

% table8---------------------------------------------------------------------
\begin{table}[t]
\centering
\caption{
   Comparison of performance on the test set of Cityscapes with state-of-the-art (trained on trainval set).
   Multi-scale and flipping testing is used for fair comparison.
}\label{table8}
\resizebox{.45\textwidth}{!}{
\begin{tabular}{c|c|c|c}
   \hline
   \hline
   Method                                           &Backbone      &Stride       &mIoU      \\
   \hline
   CCNet \cite{huang2019ccnet}                      &ResNet-101    &$8\times$    &81.90     \\
   PSPNet \cite{zhao2017pyramid}                    &ResNet-101    &$8\times$    &78.40     \\
   PSANet \cite{zhao2018psanet}                     &ResNet-101    &$8\times$    &80.10     \\
   OCNet \cite{yuan2018ocnet}                       &ResNet-101    &$8\times$    &80.10     \\
   OCRNet \cite{yuan2019object}                     &ResNet-101    &$8\times$    &81.80     \\
   DANet \cite{fu2019dual}                          &ResNet-101    &$8\times$    &81.50     \\
   ACFNet \cite{zhang2019acfnet}                    &ResNet-101    &$8\times$    &81.80     \\
   ANNet \cite{zhu2019asymmetric}                   &ResNet-101    &$8\times$    &81.30     \\
   ACNet \cite{fu2019adaptive}                      &ResNet-101    &$8\times$    &82.30     \\
   HRNet \cite{wang2020deep}                        &HRNetV2-W48   &$4\times$    &81.60     \\
   OCRNet \cite{yuan2019object}                     &HRNetV2-W48   &$4\times$    &82.40     \\
   \hline
   DeepLabV3+MCIBI (\emph{ours})                    &ResNet-50     &$8\times$    &79.90     \\
   DeepLabV3+MCIBI (\emph{ours})                    &ResNet-101    &$8\times$    &82.03     \\
   DeepLabV3+MCIBI (\emph{ours})                    &ResNeSt-101   &$8\times$    &81.59     \\
   DeepLabV3+MCIBI (\emph{ours})                    &HRNetV2-W48   &$4\times$    &\textbf{82.55}     \\
   \hline
   \hline
\end{tabular}}
\vspace{-0.5cm}
\end{table}
% table8---------------------------------------------------------------------

\noindent \textbf{Results on COCO-Stuff.}
Since previous methods only report segmentation performance trained on the outdated version of COCO-Stuff (\emph{i.e.}, COCO-Stuff-10K), our method is also only trained on this version for fair comparison.
Table \ref{table9} reports the performance comparison of our method against six competitors on the test set of COCO-Stuff-10K.
As we can see that our DeepLabV3+MCIBI adopting ResNeSt101 achieves an mIoU of $42.15\%$, outperforms previous best method (\emph{i.e.}, OCRNet leveraging HRNetV2-W48) by $1.65\%$ mIoU.
Furthermore, it is also noteworthy that our DeepLabV3+MCIBI using ResNet-101 are also superior to previous best OCRNet even when it uses a more robust backbone HRNetV2-W48.
This result firmly suggests the superiority of mining the contextual information beyond the input image.
Besides, by leveraging ViT-Large, this paper also reports a new state-of-the-art performance on the test set of COCO-Stuff-10K, \emph{i.e.}, $44.89\%$

% table9---------------------------------------------------------------------
\begin{table}[t]
\centering
\caption{
   Segmentation results on the test set of COCO-Stuff-10K.
   Multi-scale and flipping testing is employed for fair comparison.
}\label{table9}
\resizebox{.45\textwidth}{!}{
\begin{tabular}{c|c|c|c}
   \hline
   \hline
   Method                                           &Backbone      &Stride       &mIoU         \\
   \hline
   DANet \cite{fu2019dual}                          &ResNet-101    &$8\times$    &39.70        \\
   OCRNet \cite{yuan2019object}                     &ResNet-101    &$8\times$    &39.50        \\
   SVCNet \cite{ding2019semantic}                   &ResNet-101    &$8\times$    &39.60        \\
   EMANet \cite{li2019expectation}                  &ResNet-101    &$8\times$    &39.90        \\
   ACNet \cite{fu2019adaptive}                      &ResNet-101    &$8\times$    &40.10        \\
   OCRNet \cite{yuan2019object}                     &HRNetV2-W48   &$4\times$    &40.50        \\
   \hline
   DeepLabV3+MCIBI (\emph{ours})                    &ResNet-50     &$8\times$    &39.68         \\
   DeepLabV3+MCIBI (\emph{ours})                    &ResNet-101    &$8\times$    &41.49         \\
   DeepLabV3+MCIBI (\emph{ours})                    &ResNeSt-101   &$8\times$    &42.15         \\
   DeepLabV3+MCIBI (\emph{ours})                    &ViT-Large     &$16\times$   &\textbf{44.89}   \\
   \hline
   \hline
\end{tabular}}
\vspace{-0.4cm}
\end{table}
% table9---------------------------------------------------------------------

\section{Conclusion}
This paper presents an alternative perspective for addressing the context aggregation problem in semantic image segmentation.
Specifically, we propose to mine the contextual information beyond the input image to further improve the pixel representations.
First, we set up a feature memory module to store the dataset-level contextual information of various categories.
Second, a dataset-level context aggregation scheme is designed to aggregate the dataset-level representations for each pixel according to their class probability distribution.
At last, the aggregated dataset-level contextual information is leveraged to augment the original pixel representations.
Furthermore, the dataset-level representations are also used to train the classification head so that we could address intra-class compactness and inter-class dispersion from the perspective of the whole dataset.
Extensive experiments demonstrate the effectiveness of our method.

{\small
\bibliographystyle{ieee_fullname}
\bibliography{egbib}
}

\end{document}